\documentclass[preprint,12pt,authoryear]{elsarticle}
\usepackage{amssymb}
\usepackage{amsmath}
\usepackage{algorithmic}
\usepackage{graphicx, color}
\usepackage{textcomp}
\usepackage{xcolor}
\usepackage{hyperref}
\usepackage{multirow}
\usepackage{wrapfig}
\usepackage{subcaption}
\usepackage{algorithm}

\journal{Nuclear Physics B}

\begin{document}
\begin{frontmatter}

\title{Accurate and Efficient Two-Stage \\Gun Detection in Videos} 

\author{Badhan Chandra Das$^{a,b}$, M. Hadi Amini$^{a,b,}$\footnote{Corresponding authors: M. Hadi Amini and Yanzhao Wu (emails: \{moamini, yawu\}@fiu.edu).}, and Yanzhao Wu$^{a,}$\footnotemark[\value{footnote}]} 

\affiliation{organization={$^{a}$Knight Foundation School of Computing and Information Sciences, Florida International University.},
                   city={Miami},
                       state={FL},
            country={USA}}
\affiliation{organization={$^{b}$Security, Optimization, and Learning for InterDependent networks laboratory \\(solid lab)},
                       city={Miami},
                      state={FL},
            country={USA}}

\begin{abstract}
Object detection in videos plays a crucial role in advancing applications such as public safety and anomaly detection. Existing methods have explored different techniques, including CNN, deep learning, and Transformers, for object detection and video classification. However, detecting tiny objects, e.g., guns, in videos remains challenging due to their small scale and varying appearances in complex scenes. Moreover, existing video analysis models for classification or detection often perform poorly in real-world gun detection scenarios due to limited labeled video datasets for training. Thus, developing efficient methods for effectively capturing tiny object features and designing models capable of accurate gun detection in real-world videos is imperative. To address these challenges, we make three original contributions in this paper. First, we conduct an empirical study of several existing video classification and object detection methods to identify guns in videos. Our extensive analysis shows that these methods may not accurately detect guns in videos. Second, we propose a novel two-stage gun detection method. In stage 1, we train an image-augmented model to effectively classify ``Gun'' videos. To make the detection more precise and efficient, stage 2 employs an object detection model to locate the exact region of the gun within video frames for videos classified as ``Gun'' by stage 1.
Third, our experimental results demonstrate that the proposed domain-specific method achieves significant performance improvements and enhances efficiency compared with existing techniques. We also discuss challenges and future research directions in gun detection tasks in computer vision.
\end{abstract}
\end{frontmatter}
The traditional surveillance system primarily depends on human operators to analyze CCTV footage to detect such violence, which might lead to human error~\cite{HError}. Moreover, several other issues, e.g., boredom and biased profiling~\cite{surette2005thinking}, may deteriorate the effectiveness of this conventional system. In the real-world, gun detection in videos has been a highly critical task in the computer vision domain due to several factors, e.g., low video quality, small scale of gun objects, varying lighting conditions, similarity to other objects~\cite{leszczuk2024objective}, and lack of labeled video datasets to train the models effectively. Despite significant efforts, existing methods often struggle to achieve high detection accuracy, robustness, and generalize the performance in the real-world environment. The limitations of existing methods highlight the need for a more robust, accurate, and efficient solution to address the shortcomings of existing techniques in detecting guns in video footage, particularly in dynamic and challenging environments.

Previous studies have explored various Artificial Intelligence (AI)-driven approaches to address this challenge. Most of the existing video analysis models (e.g.,~\cite{tran2018closer, tong2022videomae, kondratyuk2021movinets, arnab2021vivit, fan2021multiscale}) are intended for video action recognition and have been evaluated on benchmark action recognition datasets, which may not be able to perform effective gun detection. Detection models, such as CNN~\cite{bhatti2021weapon}, R-CNN~\cite{olmos2018automatic}, faster R-CNN~\cite{alaqil2020automatic}, and different versions of YOLO~\cite{warsi2019gun, khin2024gun} are widely utilized for this task. Despite these advances, current methods may not identify guns accurately, particularly in scenarios involving low-quality video footage and unpredictable real-world conditions. Through a comprehensive empirical analysis of existing video action recognition solutions for the gun detection task, we have identified critical challenges and limitations. To address those problems, we propose a novel technique tailored towards efficient gun detection in videos. The key contributions of this work are as follows.

\begin{itemize}

    \item We empirically evaluated several state-of-the-art video action recognition models, including 3D CNN~\cite{tran2018closer}, pre-trained video models~\cite{kondratyuk2021movinets}, Transformer-based models~\cite{tong2022videomae}, hybrid architectures~\cite{xia2020lstm}, and Video FocalNet~\cite{wasim2023video}, for detecting guns in videos. Our analysis reveals that these models, despite their effectiveness in action recognition, fail to achieve high performance in gun detection.

    \item {\color{black}We propose a two-stage gun detection method for videos. First, our method combines an image-augmented video classifier, which extracts gun spatial features from video frames, with a sequence model to handle temporal correspondence. Second, to enhance efficiency and accuracy, we incorporate an object detection model that locates the exact region of the gun within video frames for the videos classified as "Gun".}

    \item {Our analysis investigates multiple application scenarios, including real-world contexts, and highlights the necessity of effective AI-driven methods for detecting guns in videos. Through empirical results, we illustrate that our proposed technique outperforms state-of-the-art methods, yielding significant improvements in both performance and efficiency for gun detection in videos.
    }

    \item Additionally, we highlight several key challenges and limitations of state-of-the-art gun detection techniques and discuss potential future research directions.
\end{itemize}
\vspace{-1ex}
\section{Motivation}
\label{motivation}

Shooting and gun violence have caused numerous fatalities and injuries almost every year in the last decade. In 2018 alone, there were more than 27 incidents of gun violence in the United States~\cite{FBI}. Apart from this, several other factors make detection more challenging.
\vspace{-2ex}
\subsection{Challenges of Human Monitored Surveillance Systems}
The efficacy of conventional surveillance systems is highly dependent on the human operator to monitor CCTV footage and respond to unusual events, such as gun violence. However, manually analyzing these videos may lead to human errors, specifically at late hours, and can cause delayed responses. Recently, several high-profile incidents of gun violence have occurred, despite the presence of CCTV surveillance and human monitoring, resulting in devastating consequences~\cite{Orlando}.

\vspace{-2ex}
\subsection{Critical Aspects of Gun Detection Models in Videos}
The primary challenge lies in extracting spatial features of such a tiny object (i.e., a gun) from videos. Furthermore, several other factors make detection even more challenging, including variability in gun size and shape, poor video quality, occlusion, obstruction, and the presence of similar objects held at unusual angles. These factors can significantly degrade model performance~\cite{leszczuk2024objective}, leading to reduced accuracy, increased false positives, and challenges in feature extraction~\cite{aqqa2019understanding}.

\vspace{-2ex}
\subsection{Lack of Domain-specific Datasets and Models}

Existing approaches, such as 3D-CNN~\cite{hara2017learning, karpathy2014large}, MoViNets~\cite{kondratyuk2021movinets}, VideoMAE~\cite{tong2022videomae}, VideoFocalNets~\cite{wasim2023video}, were primarily focused on action recognition, e.g., walking, playing, and jumping in videos. To evaluate these models, benchmark action recognition datasets or various sports classification datasets were used, e.g., the UCF101 action recognition dataset~\cite{soomro2012ucf101}, kinetics human action video dataset~\cite{kay2017kinetics}, and Sports-1M~\cite{karpathy2014large}. Therefore, there is a significant scarcity of comprehensive and labeled gun violence recognition and detection datasets within the literature. Consequently, a substantial research gap persists in developing a highly effective and efficient gun detection model, highlighting the need for further investigation in this critical area.

To combat the above challenges, we perform an empirical study to evaluate the performance of several state-of-the-art video classification methods to detect guns in videos on a synthetic gun action recognition dataset~\cite{ruiz2024firearm} and a real-world dataset~\cite{sultani2018real}. In order to address the scarcity of labeled video datasets and handle the issues of tiny object (i.e., gun) detection, we leverage the labeled gun image datasets to fine-tune the pre-trained image models with image-augmented training for extracting innate spatial gun features from the frames of video data. In addition, we improve the efficiency and robustness of our proposed method by integrating an object detection model with the classification model.

\section{Problem Statement}

In the real world, machine learning (ML) models learn through a vast amount of data in the training phase for different learning tasks. For instance, the state-of-the-art high-performing image classification models, such as ResNet~\cite{he2016deep} and EfficientNet~\cite{tan2019efficientnet}, are trained over ImageNet~\cite{deng2009imagenet}, which contains more than 1M labeled samples of 1000 categories. Using (fine-tuning) those models in another domain for any learning task also requires a significant amount of domain-specific labeled data to achieve competitive performance. As discussed earlier, existing video models are primarily focused on action recognition~\cite{kondratyuk2021movinets, tran2018closer} tasks. Again, advanced transformer-based pre-trained video recognition models, e.g., VideoMAE~\cite{tong2022videomae}, ViViT~\cite{arnab2021vivit}, Slowfast~\cite{feichtenhofer2019slowfast}, X3D~\cite{feichtenhofer2020x3d}, and MVit family~\cite{fan2021multiscale, li2022mvitv2}, which have been trained on labeled action recognition video datasets, thereby mostly suitable for video action recognition. On the other hand, object detection models are specifically designed to localize the precise region of particular objects (in this case, guns) within video frames. However, inherently, it takes longer than classification to find the desired objects in all video frames during inference. For example, if we aim to detect guns from a set of videos (some of them contain guns and others do not), the object detection model must perform inference on every frame of all videos. This exhaustive approach poses a significant limitation: it cannot indicate the absence of the desired object until all frames are fully analyzed. Consequently, it becomes a time-consuming process, particularly when the model attempts to locate a gun in videos where none are present.

In this paper, we consider gun detection in videos as a \textbf{classification-oriented object detection} problem. In this two-stage method, first, we design and train a gun video classification model to filter out instances that are negatively classified from a set of videos containing both ``Gun'' and ``No-Gun'' classes. Subsequently, we perform object detection specifically on videos that have been classified as ``Gun'' by the video classification model. This technique optimizes computational efficiency by performing object detection only on the relevant instances, reducing the processing time on videos that have already been classified as ``No-Gun''.  We formally present the method as follows.

\textbf{Stage 1: Classification model for gun videos:} Let $\mathcal{V} = \{V_1, V_2, \dots, V_N\}$ denotes a dataset with $N$ video instances, where each video $V_i$ belongs to one of two classes: $y_i \in \{\text{Gun}, \text{No-Gun}\}$. We design and train a video classification model $f_c: \mathcal{V} \to \{0, 1\}$, as
\vspace{-1ex}
\[
f_c(V_i) = 
\begin{cases} 
1, & \text{if } y_i = \text{Gun}, \\
0, & \text{if } y_i = \text{No-Gun}.
\end{cases}
\]

The classifier filters the dataset into two subsets:
\vspace{-1ex}
\[
\mathcal{V}_{\text{Gun}} = \{V_i \mid f_c(V_i) = 1\}, \quad \mathcal{V}_{\text{No-Gun}} = \{V_i \mid f_c(V_i) = 0\}.
\]

\textbf{Stage 2: Classification-oriented object detection:} For the subset $\mathcal{V}_{\text{Gun}}$, an object detection model $f_d$ is performed, where 
\vspace{-2ex}
\[
f_d: \mathcal{V}_{\text{Gun}} \to {B}^4 \times \{0, 1\}. 
\]
Here, ${B}^4$ denotes bounding box coordinates as $(x, y, w, h)$, and the output class label as a binary object detection problem. Where,
\vspace{-1ex}
\[
f_d(V_i) = \{(b_j, c_j) \mid b_j \in {B}^4, c_j \in \{0, 1\}, j = 1, \dots, M_i\},
\]
$M_i$ represents the number of detected objects in videos $V_i$, $b_j$ is the bounding box of the $j^{th}$ object, and the value of $c_j$ indicates whether a successful detection or not.

This two-stage method optimizes computational efficiency by performing object detection only on the subset $\mathcal{V}_{\text{Gun}}$, significantly reducing the time complexity compared to applying $f_d$ across the entire dataset $\mathcal{V}$.

To achieve high performance with the classification model for gun videos, labeled video datasets are required with a significant amount of samples to train a new model or fine-tune the existing models. There is a huge scarcity of labeled video datasets for such tasks in the gun video analysis domain. Therefore, current methods may not achieve high performance either in training or fine-tuning due to the scarcity of labeled training data. To tackle this problem, we obtain \textbf{transfer learning} technique that utilizes the knowledge learned from one task (\textit{source} task) of any domain and applies it to another related task (\textit{target} task) to the same or different domain~\cite{zhuang2020comprehensive}. Formally, a domain $D$ contains a feature space $X$, (X is the set of instances defined as $\{x_1, x_2,....,x_n\})$ and probability distribution $P(X)$. For a specific domain, $D = \{X, P(X)\}$. A task $\tau$ is defined as $\tau = \{Y, f\}$, here $Y$ denotes a label space defined as $\{y_1, y_2,....,y_n\}$ and $f$ is a model, $f: X \rightarrow Y$, which predicts the label $f(x)$ on a new instance $x$. The task $\tau$ is defined as $\tau = \{Y, f(x)\}$, which is learned from the training instances consisting of pairs $\{x_i, y_i\}$, where $x_i \in X$, and $y_i \in Y$. Given a source domain $D_s$ and source learning task $\tau_s$ and a target domain $D_t$ and target learning task $\tau_t$ (e.g., gun detection), where $D_s \ne D_t$ and  $\tau_s \ne \tau_t$, we aim to identify effective transfer learning strategies that can improve the learning of the target prediction function, $f_t$ in $d_t$ utilizing the knowledge in $D_s$ and $\tau_s$. 

\vspace{-1ex}
\section{Gun Detection Method}
\label{sec:method}

\subsection{Existing Video Classification Methods} 

In this paper, we explore several representative video classification models from existing research described as follows.

\noindent\textbf{3D-Convolutonal Neural Network (3D-CNN):}
3D-CNN method was proposed~\cite{tran2015learning} to capture spatial and temporal features by sliding a 3D kernel on the entire dimension of the video. 
\noindent\textbf{Mobile Video Network (MoViNet):} 
MoViNet is an efficient model for video action recognition tasks on resource-constrained devices~\cite{kondratyuk2021movinets}, which has been pre-trained on Kinetics-600 dataset~\cite{kay2017kinetics}. 

\noindent\textbf{VideoMAE:} Tong et al.~\cite{tong2022videomae} proposed video masked autoencoder (VideoMAE), a self-supervised method for pre-training. It utilizes a tube masking technique to randomly mask 90-95\% of frames and reduces computation via an asymmetric encode-decoder while training ViT backbones~\cite{dosovitskiy2020image}, effective for small datasets, e.g., UCF101~\cite{soomro2012ucf101}.

\noindent\textbf{Hybrid Architectures:} 
Hybrid architecture extracts the spatial features from the video frames with deep neural network (DNN) models (e.g., VGG) and hands the temporal features with sequence models (e.g., LSTM). Different combinations~\cite {CNN-trans},~\cite{xia2020lstm} are used for various action recognition tasks, such as gesture recognition~\cite{hu2018novel}.

\noindent\textbf{VideoFocalNet:} 
VideoFocalNet~\cite{wasim2023video} captures spatiotemporal information in videos using a two-stream focal modulation block, separately mapping spatial and temporal features and fusing them via element-wise multiplication.

However, these models are primarily focused on video action recognition tasks. Thus, their effectiveness in detecting tiny objects, such as guns in videos, particularly within real-world environments, remains uncertain.

\vspace{-2ex}
\subsection{Image-augmented Training for Video Classification}
We propose a novel gun video classification method integrated with an image-augmented training technique wrapped with a sequence model. First, we employ existing image models (e.g., VGG, ResNet, and MobileNet), where the models are pre-trained over the ImageNet benchmark dataset~\cite{deng2009imagenet}. We perform the transfer learning technique to take advantage of the pre-trained weights $(\theta_s)$ of source task $(\tau_s)$ of existing models, improving the model's ability to capture the spatial features of gun images for the target task $\tau_t$, i.e., gun detection.

Further, we apply \textbf{fine-tuning}~\cite{sharif2014cnn} technique within transfer learning, in which the parameters trained for the source task $\tau_s$, but not fully retained, rather adjusted for the target task $\tau_t$. It gets initialized by $\theta_t = \theta_s$, where $\theta_s$ and $\theta_t$ are the model parameters for source and target tasks, respectively. Then, it can be updated during training for target task $\tau_t$ as 
\begin{equation}
    \label{eqn:fine_tune}
    \theta_t = \theta_s + \Delta \theta
\end{equation}
Here, $\Delta \theta$ is the parameter update on $\theta_s$. The optimization process minimizes the loss function on the target task $\theta_t$. We leverage the model parameters $\theta_s$ trained for $\tau_s$ and fine-tune it as $\theta_t$ as shown in Equation \ref{eqn:fine_tune} for $\tau_t$, i.e., gun detection.

In order to deal with the scarcity of labeled video datasets for training gun detection models, we leverage the publicly available labeled gun image datasets collected from~\cite{olmos2018automatic} and~\cite{gu2022youtubegdd} to fine-tune the pre-trained image models, e.g., VGG, ResNet, and MobileNet. To address the challenges associated with detecting tiny objects in video frames, we employ several image augmentation techniques during fine-tuning (details provided in Section \ref{sec:Exp}). Thus, we call it an image-augmented trained model. This technique helps to carefully feed the innate spatial features to the pre-trained models during fine-tuning so that it can effectively extract the features of tiny objects (i.e., gun) from downsampled video frames. It enables the model to better generalize across diverse conditions and improves its ability to detect gun-like tiny objects even in challenging environments.

To handle the temporal features in videos, we employ sequence models (e.g., LSTM~\cite{hochreiter1997long}, GRU~\cite{chung2014empirical}, and Transformer~\cite{vaswani2017attention}) on top of the gun image-augmented classifier. We consider the videos as a sequence of frames, $X = \{x_1, x_2, ...., x_t\}$, $t$ is denoted as the length of the sequence. The input sequence $X$ is mapped to hidden state $h_t$ as $(x_t, h_{(t-1)}; \theta)$, where $\theta$ denoted as model parameters. This hidden state gets updated as $(h_{(t-1)}, x_t; \theta)$, and the inputs of the hidden state $(h_t, \theta)$, generate the output at each time step. During training, the loss function $L(Y, Y')$ is minimized according to model parameters $\theta$. Here, $Y$ and $Y'$ represent ground truth labels and predicted labels, respectively~\cite{sutskever2014sequence, goodfellow2016deep}.

\begin{figure*}[!ht]
    \centering
    \includegraphics[scale=1.1]{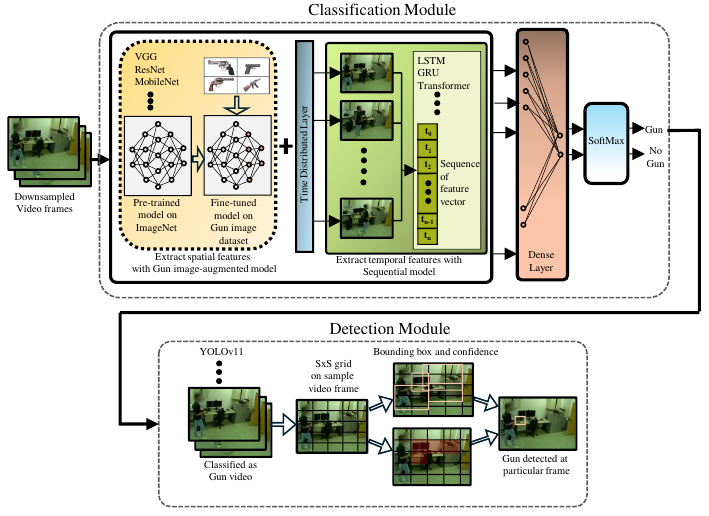}
    \caption{Overview of the proposed classification-oriented gun detection method for videos.}
    \label{fig:proposedmethod}
    \vspace{-2ex}
\end{figure*}

\subsection{Object Detection}
Unlike classification, object detection localizes the exact region of the desired object by drawing a rectangular box around each object, which is commonly known as a ``bounding box'' within the given image or video frame. You only look once (YOLO)~\cite{redmon2016you} is one of the most popular models that is frequently used for detection tasks. As its name suggests, the model is able to detect objects in video frames through a single-pass approach. Upon receiving the input video frames, first, the model divides the entire frame into a $S \times S$ grid cell, where all of these cells detect objects that fall to its center. A fixed number of bounding boxes are predicted by each grid cell with a confidence score. It also generates a class probability map, which assigns probabilities for object classes. Finally, the model applies Non-Maximum Suppression (NMS) to retain only those bounding boxes with the highest confidence score and eliminate the rest. In this paper, we fine-tune the latest version of the YOLO~\cite{khanam2024yolov11} with the gun image dataset to localize the exact region within the video frames. 

In Figure~\ref{fig:proposedmethod}, we illustrate our two-stage proposed method. Videos are provided as input to the classification module. A previously fine-tuned model (using gun image datasets with image augmentation technique) effectively extracts intrinsic spatial gun features from the downsampled video frames. A TimeDistributed layer~\cite{qiao2018time} creates a sequence of feature vectors for temporal correspondence to each of the extracted frame features. Then, the sequence model deals with the temporal features and passes them to the next layer. In the end, a softmax performs the prediction as ``Gun'' or ``No-Gun''. Next, videos classified as "Gun" proceed to the detection module, which divides frames into a $S \times S$ grid. All these grid cells predict bounding boxes, i.e., the precise location and confidence score of the ``Gun'' object class within video frames. 
\vspace{-2ex}
\subsection{Visual examples of Feature Extraction with Class Activation Map}
Class activation map is a popular visualization technique in computer vision, providing valuable insights into the feature extraction process of DNN models through visualizing the most influential spatial regions for output prediction. We use Gradient-weighted Class Activation Mapping (Grad-CAM~\cite{selvaraju2017grad}) to visualize the most significant region from where the model extracts the gun spatial features in videos.
\begin{figure}[!ht]
    \vspace{-1ex}
    \centering
    \includegraphics[scale=.35]
    {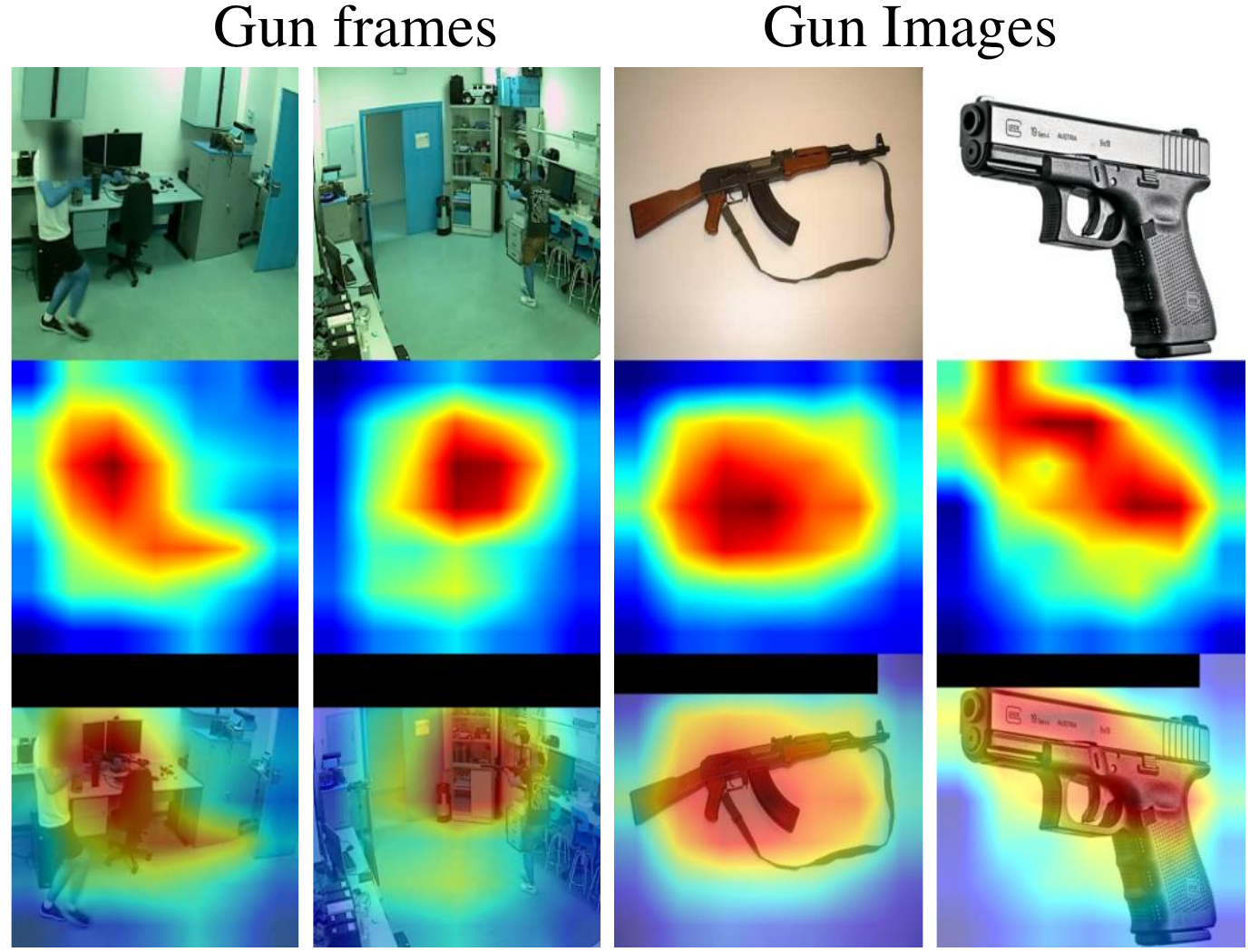}
    \caption{The class activation heatmap for downsampled video clips and gun images generated by Grad-CAM~\cite{selvaraju2017grad}}
    \label{fig:Grad-cam}
    \vspace{-1ex}
\end{figure}

To generate the Grad-CAM $L^c_{Grad-CAM}$, the output $y$ of class $c$ is generated w.r.t. feature maps $A^K$ of size $(H \times W)$ of the convolutional layer. Gradients are aggregated on the global average during backpropagation to obtain the important weights of the neurons as Equation \ref{eqn:weight}. $\alpha^{c}_K$ represents a partial liberalization of the network downsampled from $A^k$ and extracts the important weights for the target class $c$.\vspace{-1ex}

\begin{equation}
\label{eqn:weight}
    \alpha^{c}_K = \frac{1}{H \times W} \sum_{i=1}^{H} \sum_{j=1}^{W} \frac{\partial y^c}{\partial A^k_{i, j}} 
\end{equation}

The weighted combination of forward activation maps is performed, followed by $ReLU$ activation function to obtain Grad-CAM as, 
\vspace{-1ex}

\begin{equation}
    L^c_{Grad-CAM} = ReLU \sum_{K} \alpha^{c}_K A^K
    \vspace{-1ex}
\end{equation}

Figure \ref{fig:Grad-cam} illustrates the region of spatial interest where the features of the gun class have been captured for our proposed gun image-augmented fine-tuned model using the heatmap produced by Grad-Cam method \cite{selvaraju2017grad}. The two columns represent the class activation map for two representative downsampled frames with gun extracted from firearm action recognition dataset \cite{ruiz2024firearm}, and in the last two columns, we show it for gun image dataset collected from~\cite{olmos2018automatic, gu2022youtubegdd}. 

\vspace{-2ex}
\begin{table}[!h]

\centering
\caption{Dataset Information}

\scalebox{.8}{
\begin{tabular}{|c|c|c|}
\hline
\textbf{Dataset}                                                                    & \textbf{\begin{tabular}[c]{@{}c@{}}Samples per\\ Class\end{tabular}} & \textbf{Other details}\\ \hline \hline
\begin{tabular}[c]{@{}c@{}}Gun dataset for image \\ augmented training \\ \cite{olmos2018automatic} \cite{gu2022youtubegdd}\end{tabular} & \begin{tabular}[c]{@{}c@{}}Gun: 10,381\\ NoGun: 10,380\end{tabular}  & \begin{tabular}[c]{@{}c@{}}Different\\ Resolutions\end{tabular}            \\ \hline
\begin{tabular}[c]{@{}c@{}}Weapon detection datasets\\~\cite{OD} \end{tabular} & \begin{tabular}[c]{@{}c@{}}Gun Images: 2971\\ Labels: 2971\end{tabular}  & \begin{tabular}[c]{@{}c@{}}Different\\ Resolutions\end{tabular}            \\ \hline
\begin{tabular}[c]{@{}c@{}}Firearms Action\\ Recognition Dataset \\ ~\cite{ruiz2024firearm}\end{tabular}       & \begin{tabular}[c]{@{}c@{}}Gun: 303\\ NoGun: 95\end{tabular}      & \begin{tabular}[c]{@{}c@{}}Resolution: 1920x1080\\ Frame Rate: 30 FPR\end{tabular}                \\ \hline
\begin{tabular}[c]{@{}c@{}}UCF Crime Dataset~\cite{sultani2018real} \end{tabular}                         & \begin{tabular}[c]{@{}c@{}}Gun: 50\\ NoGun: 50\end{tabular}   & \begin{tabular}[c]{@{}c@{}}Resolution: 320x240\\ Frame Rate: 30 FPR\end{tabular}                  \\ \hline
\end{tabular}
}

\label{tab:dataset}
\vspace{-4ex}
\end{table}

\section{Experimental Analysis}
\label{sec:Exp}

In our experiments, first, we perform empirical studies on several existing video classification methods on the baseline action recognition dataset and the gun action recognition dataset and compare the performance. Then, we build our proposed two-stage classification-oriented gun detection model and evaluate it. This entire experiment was conducted on a GPU server with six NVIDIA A100-PCIE (40GB memory). \vspace{-1ex}
\subsection{Dataset Preparation and Evaluation Metrics}
For this experiment, we select different categories of over 10k gun images from the weapon detection datasets~\cite{olmos2018automatic, gu2022youtubegdd} for the Gun class. For the No-Gun class, we randomly pick a similar number of images (other than gun categories) from various baseline image datasets, e.g., ImageNet~\cite{deng2009imagenet}, CIFAR-10~\cite{CIFAR10}, which are commonly used in computer vision. For image augmentation, we combine several commonly used techniques in computer vision, e.g., zooming, flipping, and rotating. 
In Table \ref{tab:dataset}, we include the datasets we utilized to fine-tune both pre-trained models (e.g., VGG, ResNet, and MobileNet) and the object detection model (YOLOv11). In both empirical analysis of existing models and to evaluate our proposed model, we employ a firearm recognition and detection dataset~\cite{ruiz2024firearm} that has been specifically crafted in a controlled environment considering the factors of the CCTV footage of violent incidents, i.e., camera angle, lights, arm position in real-world gun violence video. 
Also, we utilize the UCF Crime dataset~\cite{sultani2018real}, specifically the shooting category, which consists of real-world CCTV footage from various locations, e.g., bus stops. 

First, we evaluate the gun image-augmented video classification model's performance and then show the evaluation of the classification-oriented gun detection method for the detection task. In order to evaluate the baseline methods and our proposed method, we use several widely used evaluation metrics for classification, including Accuracy (Acc.), Precision, Recall, F1 score, Area Under the Curve (AUC), and Receiver Operator Characteristic (ROC). We evaluate the detection performance of the classification-oriented gun detection method, using Average Precision (AP) since we are considering only one object class (Gun) in this experiment. Furthermore, we compare the efficiency of our proposed method with the vanilla detection technique by the execution time during inference.
\vspace{-1ex}

\begin{table}[!h]
\centering

\caption{Performances comparison of the existing models on UCF action recognition dataset~\cite{soomro2012ucf101}, firearms action recognition dataset~\cite{ruiz2024firearm}, and UCF crime dataset~\cite{sultani2018real} }
\label{tab:emprical}

\scalebox{.75}{
\vspace{-3ex}
\scalebox{.95}{
\begin{tabular}{|c|c|c|c|}
\hline
\textbf{\begin{tabular}[c]{@{}c@{}}Existing Methods\end{tabular}} & \textbf{\begin{tabular}[c]{@{}c@{}}UCF action\\recognition dataset\\(Acc.)\end{tabular}} & \textbf{\begin{tabular}[c]{@{}c@{}}Firearm action \\recognition\\  dataset (Acc.)  \end{tabular}}  & \textbf{\begin{tabular}[c]{@{}c@{}}UCF Crime\\dataset (Acc.)  \end{tabular}}\\ \hline \hline
3D-CNN~\cite{tran2018closer}                                                              & 75.31\%                                                                                  & 67.39\%               &    67.16\%                                                                   
                                                                                                                                                                                                                  \\ \hline
CNN-Transformer~\cite{CNN-trans}                                                   & 89.29\%                                                                               & 83.92\%                     &        64.93\%                                                         \\ \hline
MoViNet~\cite{kondratyuk2021movinets}                                                            & 97.94\%                                                                                  & 83.33\%             &        79.16\%                                                                 \\ \hline
VideoMAE~\cite{tong2022videomae}                                                           & 93.16\%                                                                                  & 91.34\%           &              77.63\%                                                             \\ \hline
MobileNetGRU~\cite{xia2020lstm}                                                     & 93.33\%                                                                                  & 95.04\%                               &       84.62\%                                                \\ \hline

Video-FocalNet~\cite{wasim2023video}                                                     &            97.50\%                                                                       & 83.33\%                               &        75\%                                               \\ \hline

\end{tabular}
}
}
\vspace{-2ex}
\end{table}

\subsection{Empirical Analysis of Existing Video Classifiers for Gun Detection}
In order to perform an empirical analysis, we choose five representative existing video analysis models and evaluate them on the firearm action recognition dataset~\cite{ruiz2024firearm} and UCF crime dataset~\cite{sultani2018real}, and compare their performance with the UCF human action recognition dataset~\cite{soomro2012ucf101}. In Table \ref{tab:emprical}, we observe that 3D-CNN and CNN-Transformer do not even achieve comparable performance for firearm action recognition datasets with the UCF action recognition dataset. Where the pre-trained model MoViNet gives 97.94\% accuracy on action recognition tasks, it achieves only 83.33\% for classifying gun videos. The transformer-based model, VideoMAE, achieves similar performance for gun video classification but is still less accurate than action recognition tasks. Finally, the MobileNetGRU (hybrid model) gives slightly better accuracy by 1.71\% in classifying gun videos on the manually crafted dataset in a controlled environment. Again, these methods achieve very low accuracy on the real-world UCF crime dataset, where several challenges persist as discussed in Section \ref{motivation}. The highest performance is given by MobileNetGRU, which is 84.62\%. These results underscore the necessity of high-performing and efficient models for gun detection tasks in real-world videos. 
\vspace{-2ex}

\begin{table}[!h]
\centering

\caption{Experimental results of the proposed gun image-augmented classification model for videos with different configurations for Firearm action recognition dataset~\cite{ruiz2024firearm}.}
\label{tab:firearms_results}

\scalebox{.86}{
\begin{tabular}{|c|c|c|c|c|c|}
\hline
\textbf{\begin{tabular}[c]{@{}c@{}}Method Configuration\end{tabular}}    & \textbf{Acc.}  & \textbf{Precision} & \textbf{Recall} & \textbf{F1}    & \textbf{AUC}   \\ \hline \hline
VGG + LSTM                                                                  & 93\%           & 98\%               & 89\%           & 93\%           & 96.59\%        \\ \hline
VGG + GRU                                                                   & 97\%           & 100\%              & 94\%            & 97\%           & 98.66\%        \\ \hline
\begin{tabular}[c]{@{}c@{}}VGG +  Transformer\end{tabular}                & 97\%           & 96\%               & 98\%            & 97\%           & 99.29\%        \\ \hline
ResNet + LSTM                                                               & 99\%           & 98\%              & 100\%            & 99\%           & 99.92\%        \\ \hline
\textbf{ResNet + GRU}                                                       & \textbf{100\%} & \textbf{100\%}     & \textbf{100\%}  & \textbf{100\%} & \textbf{100\%} \\ \hline
\textbf{\begin{tabular}[c]{@{}c@{}}ResNet +Transformer\end{tabular}}     & \textbf{100\%} & \textbf{100\%}     & \textbf{100\%}  & \textbf{100\%} & \textbf{100\%} \\ \hline
\begin{tabular}[c]{@{}c@{}}MobileNet +  LSTM\end{tabular}                 & 98\%           & 96\%              & 100\%            & 98\%           & 99.66\%        \\ \hline
\textbf{\begin{tabular}[c]{@{}c@{}}MobileNet + GRU\end{tabular}}          & \textbf{100\%} & \textbf{100\%}     & \textbf{100\%}  & \textbf{100\%} & \textbf{100\%} \\ \hline
\textbf{\begin{tabular}[c]{@{}c@{}}MobileNet +  Transformer\end{tabular}} & \textbf{100\%} & \textbf{100\%}     & \textbf{100\%}  & \textbf{100\%} & \textbf{100\%} \\ \hline
\end{tabular}
}
\vspace{-3ex}
\end{table}

\subsection{Evaluation of Gun Image-augmented Video Classification}

For fine-tuning the pre-trained image models with gun image datasets as mentioned in Table~\ref{tab:dataset}, we maintain 70\%, 15\%, and 15\% split ratios on the entire dataset for training, validation, and testing, respectively. We maintain the same ratio to split the firearm action recognition dataset. However, due to a lack of samples for the UCF crime dataset \cite{sultani2018real}, we divided this dataset into 50\%, 25\%, and 25\% for training, validation, and testing, respectively. To handle temporal correspondence, we utilize 512 units and 256 units for the LSTM and GRU layers, respectively, followed by dropout layers (50\% ratio) for regularization. The Transformer Encoder uses four 256 units of attention heads and 512 units of Feed-Forward Networks with a 10\% dropout rate. We train all the models for 200 epochs with early stopping criteria of no improvements of validation loss till five consecutive epochs.

We present the performance of the proposed gun image-augmented classification method in Table \ref{tab:firearms_results} for the firearm action recognition dataset in several combinations of image-augmented trained models and sequence models. It achieved significant performance improvements, and some variants outperform existing methods in terms of classification accuracy. In Table \ref{tab:firearms_results}, we show that ResNet and MobileNet combined with GRU and Transformer give a perfect 100\% accuracy across all the evaluation metrics in classifying gun videos. The values of other evaluation metrics used in this paper also show the superiority of our proposed classifier in terms of accurately detecting guns in videos for firearm action recognition datasets.

\vspace{-1ex}
\begin{table}[!h]
\centering

\caption{Experimental results of the proposed gun image-augmented classification model for videos with different configurations on UCF Crime dataset~\cite{sultani2018real}.}

\label{tab:UCFCrime_results}
\scalebox{.86}{
\begin{tabular}{|c|c|c|c|c|c|}
\hline
\textbf{\begin{tabular}[c]{@{}c@{}}Method  Configuration\end{tabular}} & \textbf{Acc.} & \textbf{Precision} & \textbf{Recall} & \textbf{F1} & \textbf{AUC} \\ \hline \hline
VGG + LSTM                                                               & 72\%          & 71\%               & 77\%            & 74\%        & 75.64\%      \\ \hline
VGG + GRU                                                                & 66\%          & 70\%               & 62\%            & 65\%        & 78.02\%      \\ \hline
\begin{tabular}[c]{@{}c@{}}VGG +  Transformer\end{tabular}             & 78\%          & 78\%               & 81\%            & 79\%        & 81.73\%      \\ \hline
ResNet + LSTM                                                            & 80\%          & 83\%               & 77\%            & 80\%        & 89.58\%      \\ \hline
ResNet + GRU                                                             & 76\%          & 79\%               & 73\%            & 76\%        & 90.86\%      \\ \hline
\begin{tabular}[c]{@{}c@{}}ResNet +Transformer\end{tabular}           & 86\%          & 81\%               & 96\%            & 88\%        & 86.38\%      \\ \hline
\begin{tabular}[c]{@{}c@{}}MobileNet +  LSTM\end{tabular}              & 90\%          & 86\%               & 96\%            & 91\%        & 97.11\%      \\ \hline
\begin{tabular}[c]{@{}c@{}}MobileNet +GRU\end{tabular}                & 84\%          & 88\%               & 81\%            & 84\%        & 92.78\%      \\ \hline
\begin{tabular}[c]{@{}c@{}}\textbf{MobileNet +  Transformer}\end{tabular}       & \textbf{92\% }         & \textbf{89\% }              & \textbf{96\%}            & \textbf{93\%}        & \textbf{98.08\%}      \\ \hline
\end{tabular}
}

\end{table}

\begin{figure*}
     \centering
     \begin{subfigure}[b]{0.32\textwidth}
         \centering
         \includegraphics[width=\textwidth, height= 1.5 in]{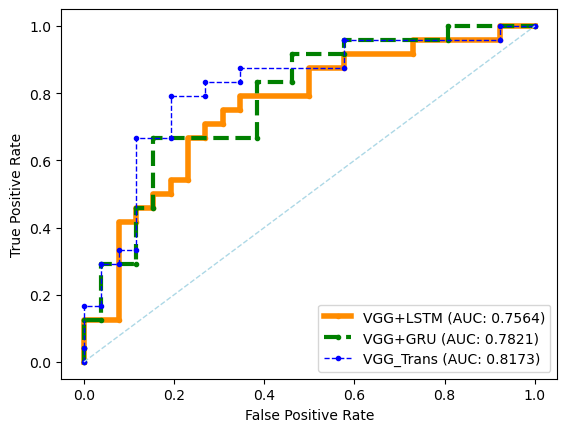}
         \caption{VGG with different sequence models}
     \end{subfigure}
     \hfill
     \begin{subfigure}[b]{0.32\textwidth}
         \centering
         \includegraphics[width=\textwidth, height= 1.5 in]{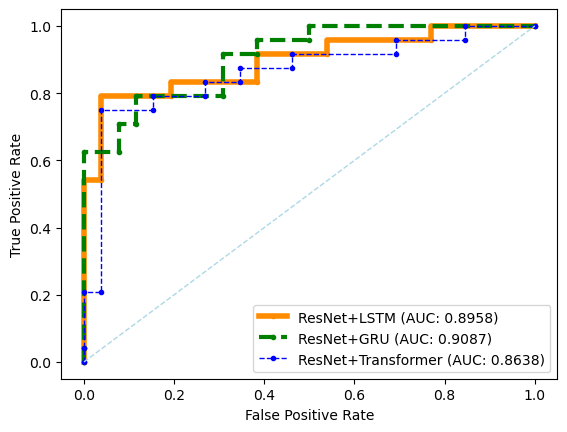}
         \caption{ResNet with different sequence models}
     \end{subfigure}
     \hfill
     \begin{subfigure}[b]{0.32\textwidth}
         \centering
         \includegraphics[width=\textwidth, height= 1.5 in]{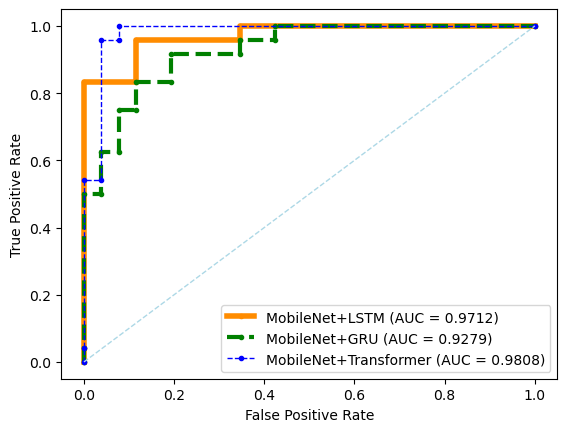}
         \caption{MobileNet with different sequence models}
     \end{subfigure}
    
        \caption{ROC curves and AUC scores of our proposed method with different configurations for UCF crime dataset.}
   
        \label{fig:ROC_UCF_crime}
      \vspace{-3ex}
\end{figure*}

\vspace{2ex}As presented in Table \ref{tab:UCFCrime_results}, we also evaluate the performance of our proposed gun image-augmented video classification model on the UCF crime dataset~\cite{sultani2018real}, which is constructed in a more real-world environment, e.g., low-light/shadow, and poor video quality. We observe that the combination of MobileNet and Transformer achieves 92\% accuracy, 89\% precision, 96\% recall, 93\% F1-score, and 98.08\% AUC score. It outperforms the highest-performing model presented in the empirical study, MobileNetGRU, with 84.62\% accuracy (see 6th row, column 4 of Table \ref{tab:emprical}). In Figure \ref{fig:ROC_UCF_crime}, we plot the ROC curves and AUC scores for all the combinations of our proposed models for the UCF crime dataset.

\vspace{-1ex}
{\color{black}

\begin{table}[!h]
{\color{black}

\caption{Performance comparison of the proposed classification-oriented Gun detection method and detection-only method on Firearm action recognition dataset~\cite{ruiz2024firearm}}
\scalebox{.72}{
\begin{tabular}{|c|c|c|c|c|c|}
\hline
\textbf{Methods}                                                                                                 & \textbf{\begin{tabular}[c]{@{}c@{}}Method\\ Configurations\end{tabular}}         & \textbf{\begin{tabular}[c]{@{}c@{}}Classification\\ Counts\end{tabular}} & \textbf{AP} & \textbf{\begin{tabular}[c]{@{}c@{}}Time\\ {[}in\\ second{]}\end{tabular}} & \textbf{\begin{tabular}[c]{@{}c@{}}Model\\ Size {[}in\\ MB{]}\end{tabular}} \\ \hline \hline
\multirow{9}{*}{\textbf{\begin{tabular}[c]{@{}c@{}}Classification-\\oriented\\ Gun Detection \\Method\end{tabular}}} & \begin{tabular}[c]{@{}c@{}}{[}VGG + LSTM{]}+\\ YOLOv11\end{tabular}              & \begin{tabular}[c]{@{}c@{}}TP: 48\\ FP: 1\\ FN: 6\\ TN: 49\end{tabular}  & 0.8413      & 185.72                                                                    & 159.35                                                                      \\ \cline{2-6} 
                                                                                                                 & \begin{tabular}[c]{@{}c@{}}{[}VGG + GRU{]}+\\ YOLOv11\end{tabular}               & \begin{tabular}[c]{@{}c@{}}TP: 51\\ FP: 0\\ FN: 3\\ TN: 50\end{tabular}  & 0.8497      & 193.57                                                                    & 158.60                                                                      \\ \cline{2-6} 
                                                                                                                 & \begin{tabular}[c]{@{}c@{}}{[}VGG + Transformer{]}+\\ YOLOv11\end{tabular}       & \begin{tabular}[c]{@{}c@{}}TP: 53\\ FP:1\\ FN: 1\\ TN: 49\end{tabular}  & 0.8552      & 185.39                                                                    & 167.371                                                                     \\ \cline{2-6} 
                                                                                                                 & \begin{tabular}[c]{@{}c@{}}{[}ResNet + LSTM{]}+\\ YOLOv11\end{tabular}           & \begin{tabular}[c]{@{}c@{}}TP: 54\\ FP: 1\\ FN: 0\\ TN: 49\end{tabular}  & 0.8570      & 234.64                                                                    & 271.82                                                                      \\ \cline{2-6} 
                                                                                                                 & \begin{tabular}[c]{@{}c@{}}{[}ResNet + GRU{]}+\\ YOLOv11\end{tabular}            & \begin{tabular}[c]{@{}c@{}}TP: 54\\ FP: 0\\ FN: 0\\ TN: 50\end{tabular}  & 0.8570      & 234.68                                                                    & 269.57                                                                      \\ \cline{2-6} 
                                                                                                                 & \begin{tabular}[c]{@{}c@{}}{[}ResNet + Transformer{]}+\\ YOLOv11\end{tabular}    & \begin{tabular}[c]{@{}c@{}}TP: 54\\ FP: 0\\ FN: 0\\ TN: 50\end{tabular}  & 0.8570      & 249.02                                                                    & 309.88                                                                      \\ \cline{2-6} 
                                                                                                                 & \begin{tabular}[c]{@{}c@{}}{[}MobileNet + LSTM{]}+\\ YOLOv11\end{tabular}        & \begin{tabular}[c]{@{}c@{}}TP: 54\\ FP: 2\\ FN: 0\\ TN: 48\end{tabular}  & 0.8570      & 203.27                                                                    & 114.83                                                                      \\ \cline{2-6} 
                                                                                                                 & \begin{tabular}[c]{@{}c@{}}{[}MobileNet + GRU{]}+\\ YOLOv11\end{tabular}         & \begin{tabular}[c]{@{}c@{}}TP: 54\\ FP: 0\\ FN: 0\\ TN: 50\end{tabular}  & 0.8570      & 199.95                                                                    & 113.33                                                                      \\ \cline{2-6} 
                                                                                                                 & \begin{tabular}[c]{@{}c@{}}{[}MobileNet + Transformer{]}+\\ YOLOv11\end{tabular} & \begin{tabular}[c]{@{}c@{}}TP: 54\\ FP: 0\\ FN: 0\\ TN: 50\end{tabular}  & 0.8570      & 214.84                                                                    & 137.87                                                                      \\ \hline
\textbf{\begin{tabular}[c]{@{}c@{}}Detection-only\\ Method\end{tabular}}                                         & YOLOv11                                                                          & N/A                                                                      & 0.8117      & 303.11                                                                    & 96.55                                                                       \\ \hline
\end{tabular}
}
\label{tab:detection_results}
}
\vspace{-2ex}
\end{table}

\subsection{Evaluation of Classification-oriented Gun Detection}
In this experiment, we used YOLOv11 for gun detection in videos. It converts the video frames to $640\times640$ size and creates grids of $80\times80$, $40\times40$, and $20\times20$ based on the stride size of 8, 16, and 32, respectively, to detect objects of varying size and scale. Predictions from all three grid scales are aggregated, ensuring robust detection across different object sizes. In Table \ref{tab:detection_results}, we illustrate the results of the classification-oriented gun detection method on firearm action recognition dataset~\cite{firearms}. Here, we observe that our proposed method performs better than the detection-only method in terms of AP values for all configurations. The detection-only method performs detection on all videos, including the No-Gun class, which may lead to several false predictions. On the other hand, our proposed classification-oriented gun detection method achieves higher AP values because, before performing detection, the classification module filters out videos that do not contain guns. Thus, the detection module yields much lower false predictions than the detection-only method. In our experiment, the proposed method yields a better AP score of 0.8570, outperforming the detection-only method by improving it by 4.53\%. Moreover, we consider the impacts of missing predictions (false negatives (FN)) yielded by our image-augmented video classifier in the proposed classification-oriented gun detection model. For example, the VGG and Transformer combination yields 1 false positive (4th row, column 3), resulting in a slight decrease in AP (0.8552), compared to the classifier combinations with no FNs (0.8570). In contrast, the VGG and LSTM combination yields 6 FNs (2nd row, column 3), leading to a higher decrease in AP. Additionally, we show the inference times for both the classification-oriented gun detection method and the detection-only method. Our observation reveals that all combinations of the proposed method take a shorter time compared to the detection-only method. Therefore, it proves the superiority of our method not only in performance but also in efficiency by moderately reducing inference time compared to the conventional detection-only method.

In Figure \ref{fig:VisualDetectionExample}, we visually present some detection examples, including correct detection, missed detection, and false prediction, along with the corresponding video length and inference time. Correct detection (1st row) demonstrates the model's ability to accurately identify and localize the region where the gun is present in the video frames. Missing detection (2nd row) highlights instances where the model fails to predict the region containing the guns in certain frames. The false prediction (3rd row) presents the cases where the model makes some mistakes in a few frames by making predictions for similar handheld objects as the gun.

\begin{figure}[!ht]

    \centering
    \includegraphics[scale=.40]{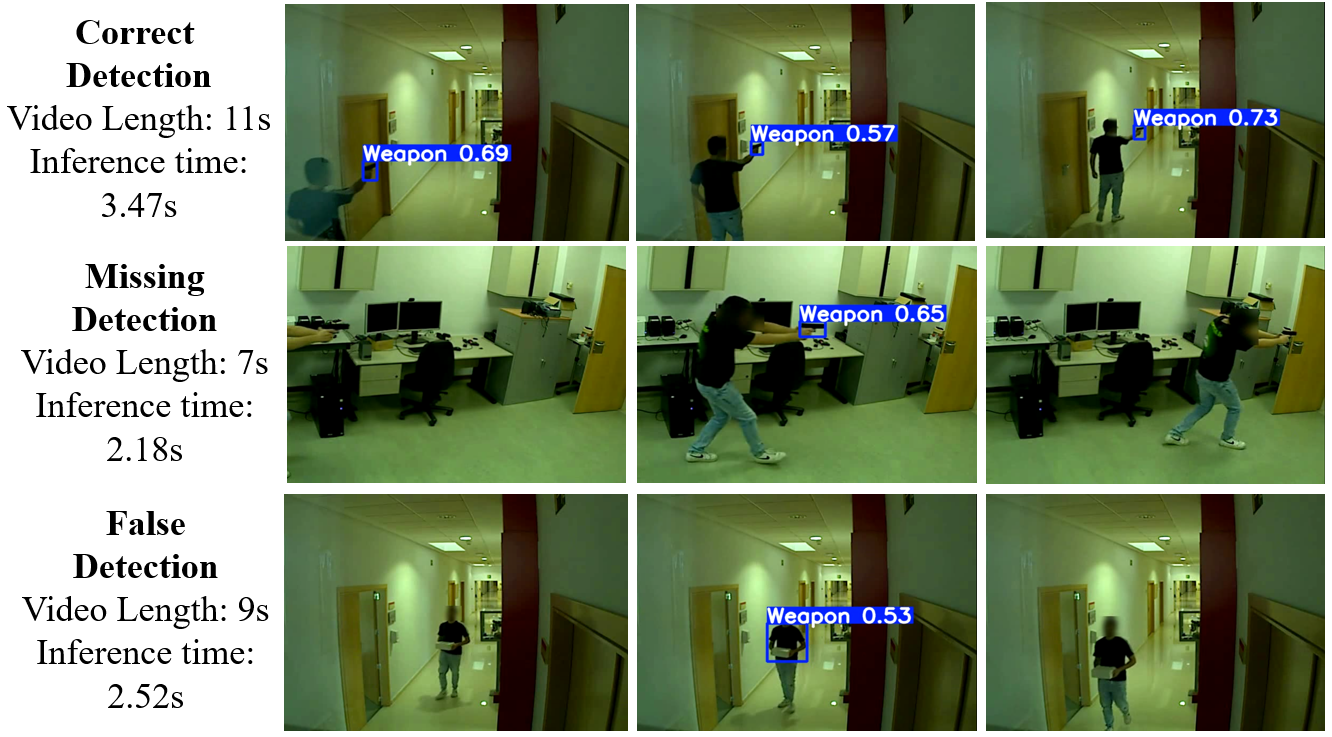}
    \caption{Visual examples of Correct, Missing, and False detection of classification-oriented gun detection method with the inference time and video length.}
    \label{fig:VisualDetectionExample}
    \vspace{-3ex}
\end{figure}

\vspace{-2ex}
\section{Discussion}

The existing models struggle to achieve high performance in detecting guns in videos due to several reasons. First, existing models are not trained specifically for identifying gun features in video frames; instead, these are trained for generic action recognition tasks~\cite{kondratyuk2021movinets, tong2022videomae}. Second, techniques such as 3D-CNN~\cite{tran2018closer}, can keep correspondence between spatiotemporal features of the action recognition tasks, however, they can not adequately capture the spatial features of guns in sequences of frames. Third, advanced methods, e.g., VideoMAE's tube masking~\cite{tong2022videomae} technique, might mask the portion where the gun appears in the video frames. Masking random cubes with extremely high ratios may cause the elimination of the spatial features of a gun where it appears in the video.
However, this idea might not work for effectively capturing tiny object features, e.g., guns.
Fourth, while several existing methods demonstrate comparable performance in action classification on synthetic datasets~\cite{ruiz2024firearm}, their effectiveness significantly diminishes in real-world scenarios. For instance, in the UCF crime dataset~\cite{sultani2018real}, where the instances are taken from real-world incidents, the performance severely deteriorates for existing models (Table~\ref{tab:emprical}, last column). Our proposed gun image-augmented video classification method addresses these issues by feeding spatial gun features to the model through fine-tuning the pre-trained models, e.g., VGG, ResNet, and MobileNet, using the image-augmentation technique. Thus, it can effectively capture the features of tiny objects, i.e., guns, in videos. Further, the sequence model handles the temporal correspondence with respect to the spatial features. According to the empirical analysis, without image-augmented training, the combination of MobileNet and GRU provides 95.04\% accuracy (i.e., 6th row  3rd column of Table~\ref{tab:emprical}) for the firearm action recognition dataset. The model facilitated with image-augmented training may achieve 100\% accuracy (i.e., $9^{th}$ row, second column of Table \ref{tab:firearms_results}) on the same dataset. Therefore, image-augmented training plays a significant role in improving the performance for classifying gun videos. Moreover, we demonstrated that our proposed gun image-augmented classifier also exhibits robust performance in real-world scenarios, e.g., the UCF crime dataset. A potential limitation of our proposed video classifier is its reduced efficacy in more challenging scenarios, such as low-quality videos, poor lighting, and inimical weather, e.g., rain or storm. Deep ensemble techniques~\cite{deep-ensembles-uncertainty, liu2019deep, heterobust} may serve as a potential solution for enhancing the overall robustness under these adverse situations.

In order to detect the precise region of the gun within the video frames, we performed object detection on those videos that have been classified as ``Gun'' by the image-augmented video classifier. Our proposed classification-oriented gun detection technique outperforms the detection-only technique in terms of both detection performance and efficiency. Furthermore, in Table \ref{tab:detection_results}, we compare the efficiency of the proposed method with the detection-only technique and observe that our proposed method enhances efficiency over the detection-only method by reducing the execution time during inference. Additionally, we include the model size of the proposed classification-oriented detection method (5th column Table \ref{tab:detection_results}) and the detection-only method (6th column Table \ref{tab:detection_results}). Our proposed method provides improved performance compared with the detection-only method in terms of both detection and efficiency; however, the model size becomes larger for all configurations than the detection-only method. The reason is that our proposed model integrates a classification module before performing detection. To deploy our proposed model for high detection performance on any resource-constrained devices, the lightest configuration, [MobileNet+GRU]+YOLOv11, can be chosen, which is 113.33MB in size.

\vspace{-1ex}
\section{Related Work}

Weapon detection in videos has been explored through various approaches in the literature. Very few of them obtained classification-based approaches, e.g., Bhatti et al.~\cite{bhatti2021weapon} explored the siding window technique through pre-trained image classifiers. Olmos et al.~\cite{olmos2018automatic} proposed a pistol detection method in videos guided by the features extracted by the VGG-16 model under the sliding window technique. On the other hand, existing detection-based approaches are also used to address this problem. R-CNN (regions with CNN features)~\cite{girshick2014rich}, faster R-CNN~\cite{ren2016faster}, different versions of YOLO~\cite{redmon2016you}, and Single Shot Detector (SSD)~\cite{liu2016ssd} are widely used in literature to detect guns or other handheld weapons (e.g., knives, grenades) in videos~\cite{ahmed2022development, hashmi2021application, bhatti2021weapon, ingle2022real}. However, these methods may not be able to achieve robust performances, specifically in more challenging real-world scenarios.

\vspace{-1ex}
\section{Conclusion}

This paper addressed the crucial challenges of gun detection in videos by contributing to three main directions. First, through an empirical analysis of existing video classification methods, we identified their drawbacks in terms of gun detection accuracy in videos. Second, to address the challenges of detecting tiny objects and limited labeled video datasets in this domain, we leveraged existing labeled image datasets to design a novel classification model for gun videos utilizing the image-augmentation technique. The method is evaluated using a wide range of datasets, including the synthetic gun action recognition dataset and real-world UCF crime dataset. Third, we proposed a classification-oriented gun detection method to efficiently localize the exact region within the video frames where the gun appears. Through experimental analysis, we have demonstrated that our proposed method not only outperforms state-of-the-art methods but also improves efficiency in detecting guns in real-world scenarios. Our findings pave the way for future research opportunities to enhance object detection capabilities and contribute to efficient gun detection.

\subsubsection*{Acknowledgments}
This material is based upon work supported by the U.S. Department of Homeland Security under Grant Award 22STESE00001-04-00. The views and conclusions contained in this document are those of the authors and should not be interpreted as necessarily representing the official policies, either expressed or implied, of the U.S. Department of Homeland Security.

\bibliographystyle{elsarticle-num-names} 
\bibliography{ref}

\end{document}